\documentclass[lettersize,journal]{IEEEtran}
\usepackage{amsmath,amsfonts}
\usepackage{algorithm}
\usepackage{array}
\usepackage[caption=false,font=normalsize,labelfont=sf,textfont=sf]{subfig}
\usepackage{textcomp}
\usepackage{stfloats}
\usepackage{url}
\usepackage{verbatim}
\usepackage{graphicx}
\usepackage{cite}
\usepackage{algpseudocode}
\usepackage{enumitem}
\usepackage{ragged2e}
\usepackage{booktabs,makecell,multirow}
\usepackage{xcolor}

\usepackage{tabularx}

\usepackage[marginal]{footmisc}

\begin{document}

\title{MG-DARTS: Multi-Granularity Differentiable Architecture Search for Trade-Off Between Model Effectiveness and Efficiency
\thanks{This manuscript has been accepted for publication in the IEEE Transactions on Neural Networks and Learning Systems (TNNLS).}
}

\author{
\IEEEauthorblockN{Xiaoyun Liu\IEEEauthorrefmark{1}, Divya Saxena\IEEEauthorrefmark{2}, Jiannong Cao\IEEEauthorrefmark{1}, Yuqing Zhao\IEEEauthorrefmark{1}, Penghui Ruan\IEEEauthorrefmark{1}\\
xiaoyun.liu@connect.polyu.hk, 
divyasaxena@iitj.ac.in,
csjcao@comp.polyu.edu.hk, csyzhao1@comp.polyu.edu.hk, penghui.ruan@connect.polyu.hk} \\
\IEEEauthorblockA{\IEEEauthorrefmark{1}Department of Computing, The Hong Kong Polytechnic University, Hong Kong
} \\
\IEEEauthorblockA{\IEEEauthorrefmark{2}School of Artificial Intelligence and Data Science, Indian Institute of Technology (IIT), Jodhpur, India}
}

\maketitle

\begin{abstract}
Neural architecture search (NAS) has gained significant traction in automating the design of neural networks. To reduce search time, differentiable architecture search (DAS) reframes the traditional paradigm of discrete candidate sampling and evaluation into a differentiable optimization over a super-net, followed by discretization. However, most existing DAS methods primarily focus on optimizing the coarse-grained operation-level topology, while neglecting finer-grained structures such as filter-level and weight-level patterns. This limits their ability to balance model performance with model size. Additionally, many methods compromise search quality to save memory during the search process. To tackle these issues, we propose Multi-Granularity Differentiable Architecture Search (MG-DARTS), a unified framework which aims to discover both effective and efficient architectures from scratch by comprehensively yet memory-efficiently exploring a multi-granularity search space. Specifically, we improve the existing DAS methods in two aspects. First, we adaptively adjust the retention ratios of searchable units across different granularity levels through adaptive pruning, which is achieved by learning granularity-specific discretization functions along with the evolving architecture. Second, we decompose the super-net optimization and discretization into multiple stages, each operating on a sub-net, and introduce progressive re-evaluation to enable re-pruning and regrowth of previous units, thereby mitigating potential bias. Extensive experiments on CIFAR-10, CIFAR-100 and ImageNet demonstrate that MG-DARTS outperforms other state-of-the-art methods in achieving a better trade-off between model accuracy and parameter efficiency. Codes are available at https://github.com/lxy12357/MG\_DARTS.
\end{abstract}


\section{Introduction}
\begin{figure}[t]
\centering
\includegraphics[width=0.85\columnwidth]{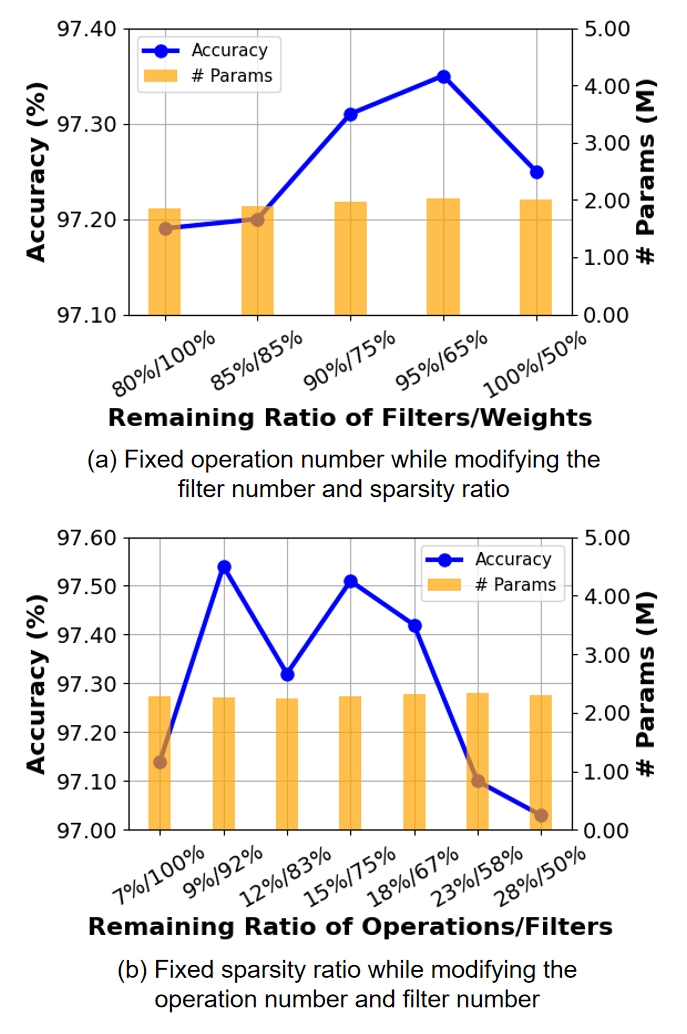} 
\caption{Illustration of the accuracy-parameter trade-off with respect to different retention ratios for units at different granularity levels (operation, weight and filters) on CIFAR-10 dataset. In (a), we maintain a constant operation number and manipulate the filter number and sparsity ratio, while ensuring the model parameters remain unchanged at approximately 2M. In (b), we keep the sparsity ratio fixed and modify the operation number and filter number, while maintaining the model parameters at around 2.3M. We observe that model accuracy varies significantly even when model sizes are similar. This underscores the necessity of effectively balancing the units of different granularities.}
\label{fig1}
\end{figure}

\IEEEPARstart{N}{eural} architecture search, the method for automatically designing effective and efficient neural networks, has gained significant popularity in various domains including computer vision \cite{liu2018darts,ma2021scenenet}, natural language processing \cite{liu2018darts,so2019evolved}. In the early stage of NAS, architectures were heuristically sampled from a search space using reinforcement learning \cite{zoph2017neural,baker2017designing} or evolutionary algorithms \cite{elsken2019efficient,salimans2017evolution}, followed by individual training and performance comparison, which incurred significant computational cost. Weight-sharing techniques were later introduced to alleviate the burden by enabling the sharing of model weights among different architectures through a super-network created from the search space. By training only a single super-net, the computational resources required can be greatly reduced. Based on that, differentiable architecture search (DAS), e.g., DARTS \cite{liu2018darts}, further improved the efficiency by introducing differentiable architecture parameters which can be optimized along with the super-net weights via gradient descent. This advancement allows for the direct updates of the architecture during super-net training, transforming the search process into the optimization and discretization of the super-net.

In spite of the great success of DAS, balancing model performance and model size remains a critical challenge in NAS \cite{ma2023pareto,wan2022e2scnet,lu2023neural}. The current research mainly focuses on coarse-grained search in constrained search spaces that only cover operation-level searchable units. While this limited granularity stabilizes the search process, it inevitably results in the presence of redundant parameters in the discovered architectures, which consequently affects model efficiency. 

To eliminate the redundancy, recent studies have emerged to expand the search space and jointly investigate finer-grained units at the filter and weight levels. While certain studies explore filter-level search \cite{stamoulis2019single,fu2020autogan,wan2020fbnetv2, sukthanker2024weight, peng2024recnas} and weight-level search \cite{you2022supertickets,mousavi2023dass}, most of them overlook the coordination of retention ratios of units at different granularities. However, the retention ratio itself plays a critical role in determining model performance. For example, it is mentioned in \cite{jin2022pruning} that there exists an optimal sparsity ratio for a specific neural network in terms of achieving the best performance. We further observe that such performance discrepancies cannot be solely attributed to the differences in model sizes introduced by sparsity. Even models with similar parameter counts can exhibit great performance variations depending on different retention ratios of the operations, filters and weights, as shown in Fig. \ref{fig1}. This underscores the importance of adaptively adjusting unit proportions across different granularity levels. Failing to consider this may result in excessive pruning at certain levels, particularly in the absence of full super-net pretraining, ultimately degrading model performance.

Additionally, DAS requires the whole super-net to be placed in the memory for optimization and discretization, leading to substantial memory usage. This issue becomes more pronounced when the finer-grained searchable units are introduced. For example, in order to facilitate the flexible filter-level search, the initial super-net must possess a larger number of filters for candidate selection, which increases the demands of the memory. Several techniques have been proposed to reduce memory consumption during search. Path-level decoupling strategies \cite{cai2019proxylessnas,dong2019searching,hou2024decoupled, huang2023u, kim2023device, wang2023fp} activate only a limited subset of operations at each iteration, but often overlook inter-path interactions, limiting the discovery of high-performing multi-path architectures. Depth-level \cite{liu2018darts,yu2022cyclic,li2020block,li2021bossnas,liu2018progressive,mousavi2023dass, hao2025mig} and width-level \cite{xu2020pcdarts,xue2022partial,yin2025spatial,wei2025csc,cai2023epc,li2024adaptive,hu2024l,xue2023improved,hao2025mig} strategies simplify or decompose the network structure during search, but may introduce bias, as the searchable units are prematurely evaluated without considering the overall architecture. As a result, there is a necessity to propose a more general and effective memory reduction strategy applicable to flexible and complex search spaces. 

To address these issues, we propose Multi-Granularity Differentiable Architecture Search (MG-DARTS), a unified framework that enables comprehensive and memory-efficient exploration of a multi-granularity search space, aiming to achieve a better trade-off between model performance and size. Specifically, we introduce a granularity-aware adaptive pruning mechanism that jointly prunes operations, filters, and weights within a unified framework. This is achieved by learning discretization functions specific to each granularity level, allowing the pruning criteria to adapt to the evolving architecture. As a result, MG-DARTS can flexibly determine appropriate retention ratios across different granularities to meet various target model sizes. To reduce memory consumption, we further propose a multi-stage search strategy that decomposes super-net optimization and discretization into a sequence of sub-net stages. Each stage operates on a reduced sub-net, maintaining low memory usage. However, early-stage pruning may introduce bias, as it is performed without foresight of subsequent architectural choices. To address this, we introduce progressive re-evaluation, allowing previously pruned units to be reconsidered and previously retained units to be further pruned, thereby mitigating bias across stages.

Through extensive experiments, our method demonstrates promising outcomes in various aspects. \textbf{First}, it reduces the model size without compromising the performance compared with coarse-grained methods. \textbf{Second}, it enhances the model performance while maintaining a similar model size compared with existing fine-grained methods. \textbf{Third}, it saves memory during the search process without diminishing the quality of the search.

The main contributions of this work can be summarized as follows:
\begin{enumerate}
\item To the best of our knowledge, we conduct the first in-depth study on adaptively adjusting the retention ratio at different granularity levels in NAS to trade off between model effectiveness and model efficiency. 
\item We propose an effective memory reduction strategy that can be applied to the expanded DARTS search spaces without compromising the search quality.
\item We demonstrate the superiority of MG-DARTS through extensive experiments on CIFAR-10, CIFAR-100 and ImageNet. Compared with baseline methods, MG-DARTS achieves a better trade-off between model performance and model size.
\end{enumerate}

\section{Related works}
In this section, we will first introduce the recent progress in the search space expansion in DAS. Then, we summarize widely employed memory reduction strategies.

\subsection{Search Space Expansion}
Considering search efficiency, most DAS approaches adopt cell-based search spaces \cite{elsken2019neural} such as the widely used DARTS search space. Recent works have attempted to relax the constraints within these search spaces to enable the discovery of more diverse and effective network designs.

To enable more flexible network topologies, Amended-DARTS \cite{bi2019stabilizing} explores independent cell structures. GOLD-NAS \cite{bi2020gold} further allows each node to preserve an arbitrary number of precedents and each edge to preserve an arbitrary number of operations. 

Motivated by pruning techniques, other works turn to introduce finer-grained searchable units. At the filter level, FBNetV2 \cite{wan2020fbnetv2} defines a set of candidate blocks with varying filter numbers for each layer and formulates filter number selection as a differentiable process. Single-Path NAS \cite{stamoulis2019single} encodes all candidate convolution operations into a single `superfilter' and jointly searches for the optimal kernel size and expansion ratio by applying learnable thresholds over different subsets. TangleNAS \cite{sukthanker2024weight} enhances this approach by introducing additional architecture parameters to better capture the importance of each subset. ReCNAS \cite{peng2024recnas} proposes dynamic programming-based channel pruning to enable more fine-grained filter pattern selection. Beyond filter-level granularity, weight-level search has also been explored. For example, SuperTickets \cite{you2022supertickets} iteratively prunes unimportant operations and weights to identify the efficient architectures and their `lottery sub-networks' in an end-to-end way. DASS \cite{mousavi2023dass} incorporates two sparse candidate operations to promote sparsity-friendly network discovery. 

However, most of these methods focus primarily on optimizing unit distribution within individual granularity levels, while ignoring the allocation trade-offs across different levels of granularity. 

In this work, we follow GOLD-NAS to relax the constraints on network topology to enable variation in the final operation number within the architecture. Building on this flexibility, we incorporate operation-level, filter-level, and weight-level units to construct a unified multi-granularity search space. On top of this, we propose a general differentiable architecture search framework to jointly optimize the retention ratios of units across different granularities, with the goal of discovering architectures that achieve a better trade-off between accuracy and parameter efficiency. 

\subsection{Memory Reduction Strategies}
Some works adopt path-level decoupling strategies, where only a subset of operations is activated at each iteration. ProxylessNAS \cite{cai2019proxylessnas} proposes a path binarization strategy to activate only one or two paths in the super-net. GDAS \cite{dong2019searching} and D2NAS \cite{hou2024decoupled} employ the Gumbel-Max trick for discrete path sampling. On-NAS \cite{kim2023device} adopts an expectation-based sampling approach over both edge pairs and operations. U-DARTS \cite{huang2023u} applies random sampling. FP-DARTS \cite{wang2023fp} partitions candidate operations into two disjoint subsets and treats each subset as a unified path. However, these approaches are limited in discovering well-performing multi-path architectures due to biased optimization that neglects the interaction among operations on the same edge.

In contrast, depth-level proxy or decomposition strategies aim to reduce search cost by operating on shallower networks. DARTS \cite{liu2018darts} and its subsequent works \cite{yu2022cyclic,mousavi2023dass} search over a shallow super-net with repeatable cells and later stack them to form a deeper target network. To fill the performance gap caused by this proxy-target mismatch, PDARTS \cite{chen2019progressive}, PNAS \cite{liu2018progressive}, and MIG-DARTS \cite{hao2025mig} progressively reduce the number of candidate operations while increasing the network depth. DNA \cite{li2020block} factorizes the super-net along the depth axis into blocks and applies block-wise supervision with a pre-trained teacher model. BossNAS \cite{li2021bossnas} instead leverages self-supervised learning for block-wise guidance. 

Similarly, width-level proxy strategies improve efficiency by reducing the number of active channels during search. PC-DARTS \cite{xu2020pcdarts} and its subsequent works \cite{yin2025spatial,wei2025csc} samples a subset of channels for operation selection while bypassing the held out part in a shortcut. EPC-DARTS \cite{cai2023epc} and ACA-PC-DARTS \cite{li2024adaptive} enhance this by introducing attention modules for channel selection. $l$-DARTS \cite{hu2024l} reparameterizes mixed operation outputs through weighted normalization to improve stability. PA-DARTS \cite{xue2023improved} and MIG-DARTS \cite{hao2025mig} further apply progressive channel scaling to alleviate the performance gap.

However, both depth-level and width-level strategies may introduce bias, as the importance of previously evaluated units may vary in the full-scale architecture. 

There are also some works which directly limits the number of candidate operations \cite{bi2019stabilizing,bi2020gold}, starts from a small set \cite{guo2020breaking}, or explore more parameter-efficient candidate operations \cite{guo2021towards}. 

In this work, we propose multi-stage search, a depth-level memory reduction strategy augmented with a re-evaluation mechanism to mitigate the aforementioned bias. 

\section{Preliminaries}
In this section, we present the frequently studied single-granularity DARTS search space, aiming to establish a foundational understanding in preparation for the introduction of the multi-granularity search space in the next section.

\subsubsection{DARTS Search Space}
The original DARTS search space consists of multiple cells, wherein each cell is a directed acyclic graph encompassing an ordered sequence of $M$ computational nodes. Each node $x_i$ serves as a latent representation and each directed edge $(i,j)$ denotes the information transformation from $x_i$ to $x_j$. The operations $o$ on each edge need to be determined, where the operation types include convolutions with different filter sizes, pooling and skip connection.

During the search, DARTS relaxes the discrete search space to be continuous by constructing a super-net incorporating all the operations and combining them utilizing differentiable architecture parameters. Within the super-net, the output of each node is the weighted sum of all its precedents. Mathematically, there is $x_j=\sum_{i<j}\sum_{o \in O} \sigma(\alpha_{i,j}^o) \cdot o(x_i)$, where $O$ indicates the candidate operation set, $o$ represents a candidate operation, $\alpha_{i,j}^o$ denotes the architecture parameter of the operation $o$ on the edge $(i,j)$, and $\sigma(\alpha_{i,j}^o)$ signifies the scaled architecture parameter. In this way, the problem of selecting the operations $o$ is transformed into that of optimizing and discretizing the scaled architecture parameters $\sigma(\alpha)$. 

\begin{figure}[t]
\centering
\includegraphics[width=0.85\columnwidth]{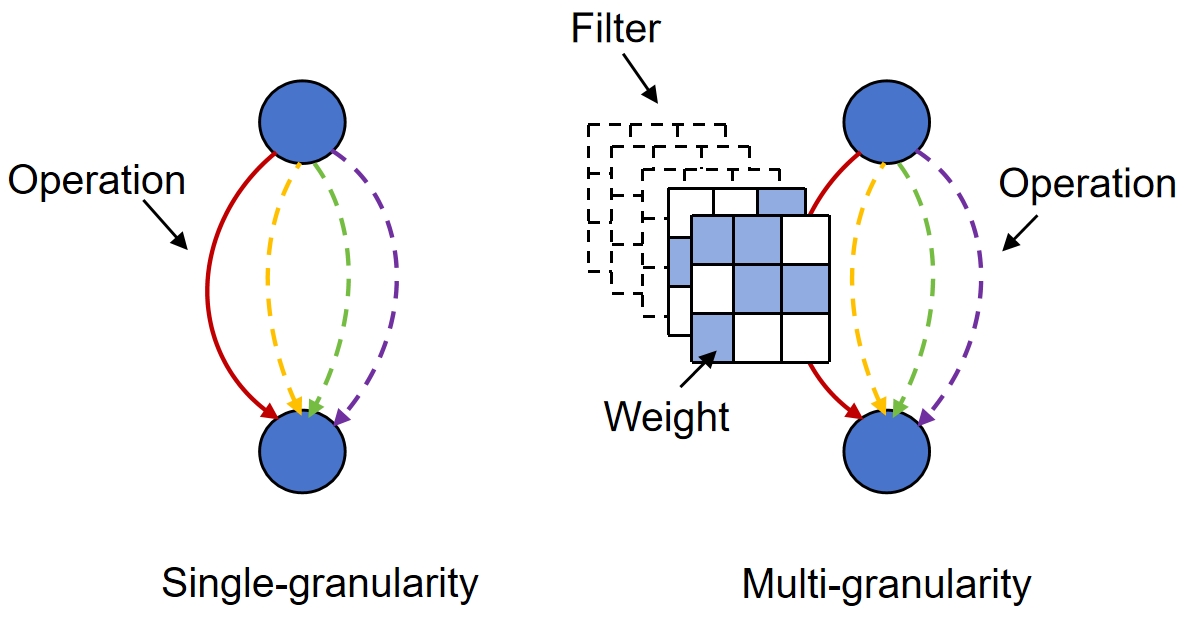} 
\caption{Illustration of the single-granularity and multi-granularity search space. The multi-granularity search space allows for the exploration of fine-grained filter-level and weight-level units, which enables a greater reduction of potential redundant parameters and facilitates the discovery of more light-weight yet effective models.}
\label{fig3}
\end{figure}

\section{Methodology}
In this section, we first describe the multi-granularity search space to be explored. Then, we introduce our proposed MG-DARTS framework, which consists of 1) adaptive pruning to balance the retention ratios of units at different granularity levels in the discovered networks, and 2) multi-stage search to reduce the memory usage without compromising the search quality.

\begin{figure*}[t]
\centering
\includegraphics[width=1\textwidth]{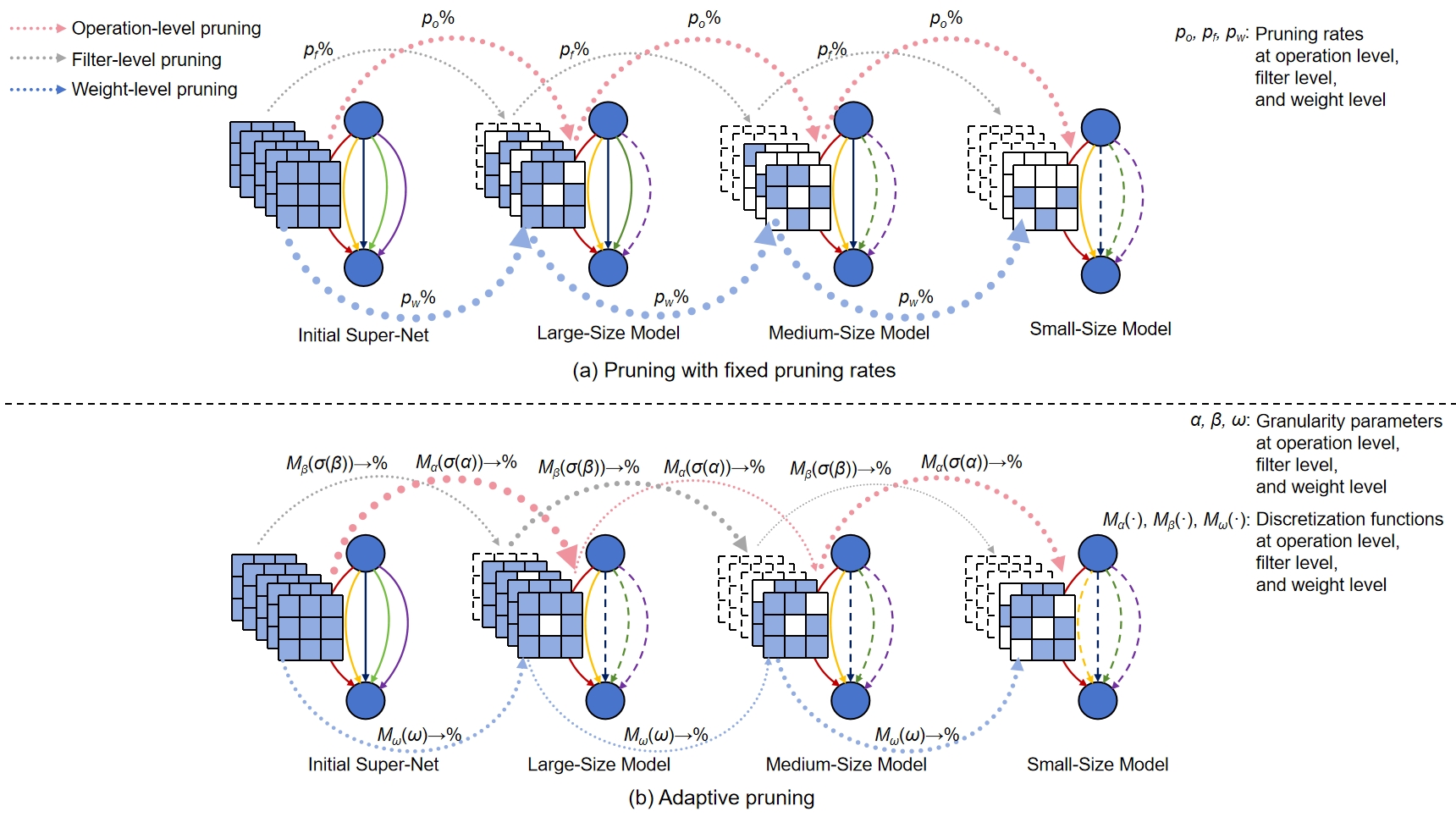} 
\caption{Overview of pruning with fixed pruning rates and adaptive pruning. The dotted arrows with different colours represent the pruning process at different granularity levels. The width of the dotted arrow indicates the number of the pruned units. Existing works independently prune on different granularities with fixed pruning rates. In contrast, our adaptive pruning learns granularity-specific discretization functions to adaptively determine the pruning ratio at each granularity level according to the architecture evolution, so that the retention ratios at different levels can be optimized for models with different sizes.}
\label{fig2_2}
\end{figure*}

\subsection{Multi-Granularity Search Space}
Compared with the single-granularity search space mentioned in the previous section, the multi-granularity search space covers not only operation-level but also filter-level and weight-level searchable units, as illustrated in Fig. \ref{fig3}. Specifically, apart from the operation type, we also need to determine the filter number and sparsity pattern in convolution operations. 

The problem of searching for the proper filter number and sparsity pattern can be viewed as that of selecting the important subset of filters and weights from the super-net. This naturally unifies the search across different granularity levels into the same subset selection paradigm used in traditional NAS. The differentiable architecture parameter $\alpha$ can be viewed as the operation-level granularity parameter that represent the operation importance. Inspired by this, we introduce differentiable filter-level granularity parameter denoted as $\beta$ to indicate the significance of each filter. For an individual weight, we directly use the magnitude of the weight itself $\omega$ to represent its importance. Mathematically, in each convolution operation $o(x_i)$ in the constructed super-net, there is $C_{out}^f=\sigma(\beta^f) \cdot \text{Conv}(C_{\text{in}}, \omega_f)$, where $C_{in}$ is the input feature map, $f$ refers to a filter, $C_{out}^f$ is the feature map generated by $f$, $\omega^f$ denotes the weights in the filter, $\beta^f$ represents the granularity parameter of $f$, and $\sigma(\beta^f)$ is the scaled granularity parameter. In this super-net, we need to optimize and discretize the scaled granularity parameters of the three levels, i.e., $\sigma(\alpha)$, $\sigma(\beta)$ and $\omega$.

\subsection{Adaptive Pruning}
As discussed above, the primary concern in exploring the multi-granularity search space is the joint optimization and discretization of the scaled granularity parameters $\sigma(\alpha)$, $\sigma(\beta)$ and $\omega$. Previous studies \cite{you2022supertickets,mousavi2023dass} have concentrated on improving joint optimization to accurately compare the unit importance within each granularity level. They achieve this by adaptively updating the granularity parameters of different levels, taking into account the changes in unit importance caused by the evolution of units at other levels. However, they independently discretize the granularity parameters at different levels with fixed pruning rates, failing to optimize the retention ratios of multi-level units. In this case, we focus on improving joint discretization with adaptive pruning to adaptively determine the retention ratios according to the evolving architecture. To achieve this, we need to enable the comparison of unit importance across different granularities. Given that the granularity parameters are measured on different scales, a direct numerical comparison is not feasible. Therefore, we propose the dynamic acquisition of granularity-specific pruning criteria to determine when a unit can be considered unimportant within its own granularity level along with the architecture evolution, enabling implicit comparison of unit importance across granularities.

Technically, instead of manually defining pruning rates or importance thresholds for discretization, we learn granularity-specific discretization functions $M_\alpha(\cdot), M_\beta(\cdot), M_\omega(\cdot)$ with differentiable thresholds $t_\alpha,t_\beta,t_\omega$ during the super-net optimization, as shown in Fig. \ref{fig2_2}. The objective of the discretization functions is to zero out the scaled granularity parameters $\sigma(\alpha),\sigma(\beta),\omega$ whose magnitudes are small enough. We formulate them as
\begin{equation}
    M_\alpha(\sigma(\alpha))=S(|\sigma(\alpha)|-t_\alpha) \cdot \sigma(\alpha), \label{eq1_1}
\end{equation}
\begin{equation}
    M_\beta(\sigma(\beta))=S(|\sigma(\beta)|-t_\beta) \cdot \sigma(\beta), \label{eq1_2}
\end{equation}
\begin{equation}
    M_\omega(\omega)=S(|\omega|-t_\omega) \cdot \omega, \label{eq1_3}
\end{equation}
where $S(\cdot)$ is a binary step function, $t_\alpha,t_\beta,t_\omega$ are trainable thresholds. To make $S(\cdot)$ differentiable, we follow \cite{dettmers2019sparse} to adopt a long-tailed higher-order estimator.

Considering the operation differences, we further refine $M_\alpha(\cdot),M_\beta(\cdot),M_\omega(\cdot)$ at the operation level to $M_\alpha^o(\cdot),M_\beta^o(\cdot),M_\omega^o(\cdot)$ in practice. Note that we have $M_\alpha^o(\cdot)$ for all kinds of operations, but only have $M_\beta^o(\cdot)$ and $M_\omega^o(\cdot)$ for convolution operations. Applied in the forward propagation of the super-net, there is
\begin{equation}
    x_j=\sum_{i<j}\sum_{o \in O} M_\alpha^o(\sigma(\alpha_{i,j}^o)) \cdot o(x_i), \label{eq2}
\end{equation}
and in each convolution operation $o$, there is
\begin{equation}
    C_{out}^f=M_\beta^o(\sigma(\beta^f)) \cdot \text{Conv}(C_{in}, M_\omega^o(\omega^f)). \label{eq3}
\end{equation}

By learning the discretization functions $M_\alpha(\cdot),M_\beta(\cdot),M_\omega(\cdot)$, we obtain tailored pruning criteria for each granularity adaptively according to the evolving architecture, so that the unit retention ratios can be properly determined. 

For a more stable search process, we follow the gradual pruning process proposed in \cite{bi2020gold}. This procedure involves conducting the discretization gradually along with the super-net optimization with regularization of resource efficiency (e.g., FLOPs), rather than all at once at the end. We prune out the units whose updated scaled granularity parameters $\sigma(\alpha),\sigma(\beta)$ or $\omega$ computed by $M_\alpha(\cdot),M_\beta(\cdot)$ or $M_\omega(\cdot)$ become zero after $e$ epochs. The overall adaptive pruning scheme is summarized in Algorithm \ref{alg1}.

At the operation level, the pruned operations are replaced with a non-parametric zero operation, which turns the input into zero. Considering the presence of skip connection operations, we align the number of filters in the output layer of each convolution operation within the same cell to match the average number of output filters among all convolution operations in the cell.  

\begin{algorithm}[tb]
\caption{Adaptive Pruning}\label{alg1}
\textbf{Input}: dataset $D$, target model size $Param_{min}$, initial remaining units $U_{act}$\\
\textbf{Parameter}: granularity parameters $\alpha, \beta, \omega$, thresholds $t_{\alpha},t_{\beta},t_{\omega}$\\
\textbf{Output}: an architecture consisting of the remaining units $U_{act}$
\begin{algorithmic}[1] 
\State $epoch \gets 0$
\While{$Param(U_{act}) \geq Param_{min}$}
\State $epoch \gets epoch+1$
\State Update loss $L \gets Loss(D;\alpha^{U_{act}},\beta^{U_{act}},\omega^{U_{act}})$ \Comment{Optimization}
\State Update $\alpha^{U_{act}}, \beta^{U_{act}}, \omega^{U_{act}}$ by gradient descent based on $L$ 
\State Update $t_{\alpha}, t_{\beta}, t_{\omega}$ by gradient descent based on $L$
\For {each granularity param $g$ in $\alpha, \beta, \omega$} \Comment{Discretization}
\State Update $\sigma(g)$ with discretization function $M_g(\cdot)$: $\sigma(g) \gets S(|\sigma(g)|-t_g) \cdot \sigma(g)$ 
\For {unit $u$ in $U_{act}$} 
\If {$\sigma(g^u)=0$}
\State Update remaining units $U_{act} \gets U_{act} \setminus u$
\EndIf
\EndFor
\EndFor
\EndWhile
\end{algorithmic}
\end{algorithm}

\subsection{Multi-Stage Search}
\begin{figure*}[t]
\centering
\includegraphics[width=0.87\textwidth]{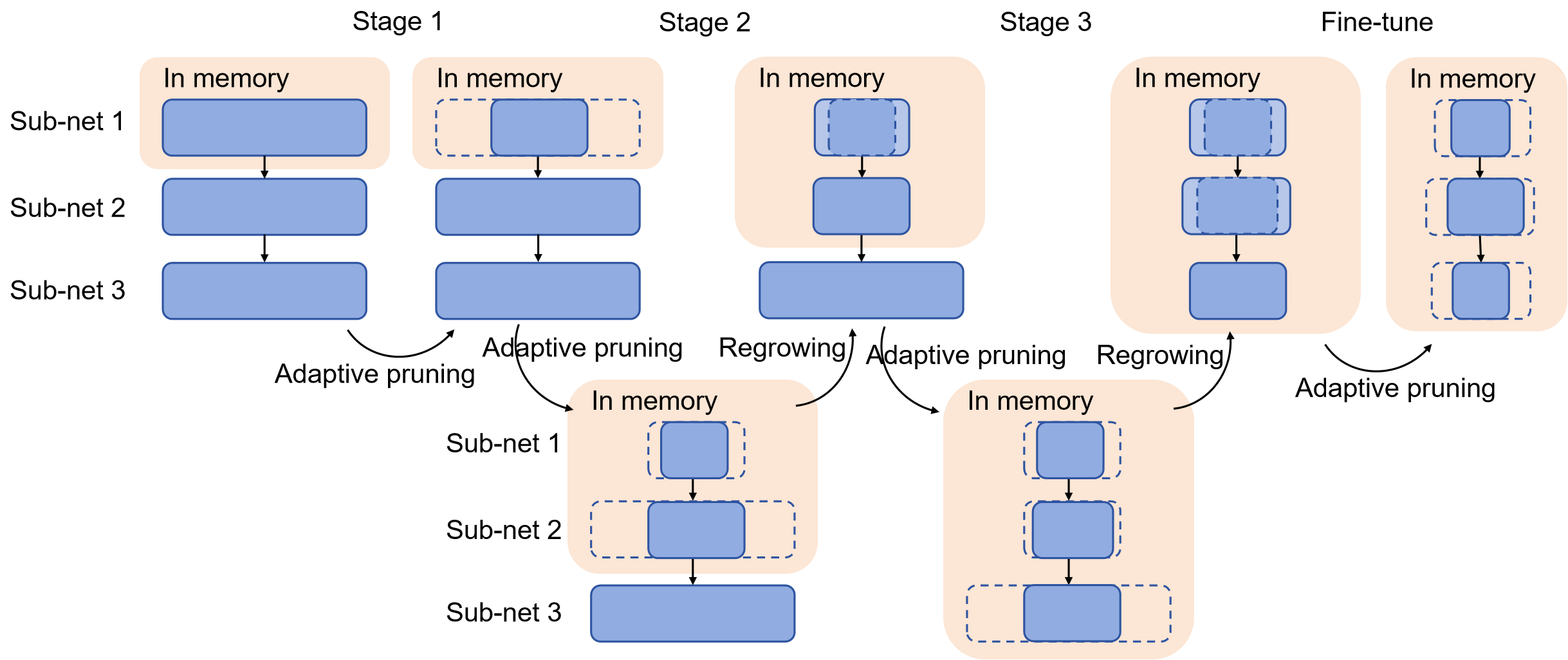} 
\caption{Illustration of multi-stage search. We decompose the super-net optimization and discretization into multiple sub-net stages to save memory, and enable further pruning and regrowing of the units in previous sub-nets during subsequent stages to reduce bias.}
\label{fig2}
\end{figure*}

The high memory consumption of DAS comes from the large number of parameters in the super-net and their corresponding gradients. We maintain low memory consumption by breaking down the super-net optimization and discretization into multiple sub-net stages. Different from previous decomposition-based methods, we enable re-evaluation of the remaining and pruned units to compensate for the potential degradation of search quality caused by bias, as illustrated in Fig. \ref{fig2}. 

Let the search space of the whole architecture be N. To realize the multi-stage search, we decompose $N$ into sequential n moderate-size search spaces {$N_1,...,N_i,...,N_n$} along the network depth, where $N_{i+1}$ is originally connected with $N_i$. In this way, the super-net constructed from the search space is naturally separated into n sub-nets with much smaller sizes. We first coarsely search for the sub-net structures in various stages, removing most of the unimportant units, and then fine-tune the remaining architecture until the target model size is achieved. 

\begin{algorithm}[tb]
\caption{Multi-Stage Search}\label{alg2}
\textbf{Input}: dataset $D$, target model size $Param_{min}$, sub-net number $n$, target sub-net size $Param_{min}^{s}$ ($s=0,1,...,n$), regrowing ratio $r$, units in the search space $U$\\
\textbf{Parameter}: scaled granularity parameters ${\sigma(\alpha), \sigma(\beta)}$, ${\omega}$, pruned operation number $num_p$, remaining units $U_{act}$\\
\textbf{Output}: an architecture consisting of the remaining units $U_{act}$
\begin{algorithmic}[1] 
\State Initialize remaining units $U \gets \emptyset$
\For {stage $s=0,1,...,n$}
\State Initialize pruned operation number $num_p \gets 0$
\State Update remaining units $U_{act} \gets U_{act}\cup {U^{sub-net_s}}$ 
\While{$Param(U_{act}) \geq Param_{min}^{s}$} \Comment{Pruning}
\State Adaptive Pruning on $U_{act}$
\State Update pruned operation number $num_p \gets num_p + num^{prune}$
\EndWhile
\For {each cell $c$ in $U^{act}$} \Comment{Regrowing}
\For {each operation type $T$ in $c$}
\State Calculate regrowing potential $p^T_{c}$
\State Calculate regrowing number ${num_r}^{T}_{c} \gets num_p\times r \times p^T_{c}$
\State Regrow ${num_r}^{T}_{c}$ operations of type $T$ in cell $c$ 
\State Update remaining units $U_{act} \gets U_{act} \cup U^{regrow}$
\EndFor
\EndFor
\EndFor
\While{$Param(U_{act}) \geq Param_{min}$} \Comment{Fine-tune}
\State Adaptive Pruning on $U_{act}$
\EndWhile
\end{algorithmic}
\end{algorithm}

To supervise the optimization of the sub-net, we add task-specific output layers with predefined structure on top of each sub-net to generate results, so that the ground truth can be directly used as the supervision signal. However, since the task-specific output layers differ from the actual high-level layers in the target network, there is bias when optimizing previous sub-nets. Specifically, due to the complex interconnections between different units, the importance of the same unit in previous sub-nets may greatly change with the evolution of the following sub-nets. 

To mitigate the bias, we perform progressive re-evaluation of units in preceeding sub-nets. On one hand, we continue to update the discretization functions for preceeding sub-nets during subsequent stages to further prune unimportant units. 

On the other hand, we estimate the potential of the pruned units for regrowing. Since the regrowth of operations will reinstate filters and weights, we mainly focus on the operation-level regrowing. Intuitively, if certain operations in a cell have a substantial impact on network performance at a given moment, reinserting the same operations in that cell should prove beneficial. Thus, we assess the regrowing potential for each operation type individually across different cells. 

Inspired by \cite{dettmers2019sparse}, we utilize both the mean magnitude and the mean momentum of the granularity parameters $\alpha$ of each operation type in each cell to indicate the potential. The magnitude measures the direct contribution of the operation to accuracy, while the momentum gauges the persistent error reduction ability of the operation through the exponentially smoothed gradients. The momentum is formulated as 
\begin{equation}
    mom(\sigma(\alpha))^{t}=\gamma \cdot mom(\sigma(\alpha))^{t-1}+(1-\gamma) \cdot \frac{\partial L}{\partial \sigma(\alpha)}^{t-1}, \label{eq4}
\end{equation}
where $\gamma$ is a smoothing factor and $L$ is the loss. Therefore, the potential of the operation type $T$ in cell $c$, $p_{c}^{T}$, can be calculated as following:
\begin{equation}
    mean\_mag_{c}^{T}=\frac{\sum_{j<M}\sum_{i<j} \sigma(\alpha_{c_{i,j}}^{T})}{\sum_{j<M}\sum_{i<j}S(\sigma(\alpha_{c_{i,j}}^{T}))}, \label{eq5}
\end{equation}
\begin{equation}
    mean\_mom_{c}^{T}=\frac{\sum_{j<M}\sum_{i<j} mom(\sigma(\alpha_{c_{i,j}}^{T}))}{\sum_{j<M}\sum_{i<j} S(\sigma(\alpha_{c_{i,j}}^{T}))}, \label{eq6}
\end{equation}
\begin{equation}
    p_{c}^{T} = \frac{1}{2}(mean\_mag_{c}^{T}+mean\_mom_{c}^{T}). \label{eq7}
\end{equation}

We set one hyper-parameter, the regrowing ratio $r$, to control the regrowth number $num_r$:
\begin{equation}
    num_{r} = r \cdot num_{p} \label{eq8},
\end{equation}
where $num_{p}$ is the number of pruned operations.

Based on the calculated regrowing potential, we allocate the regrowing quota to different operation types in different cells. Within each operation type in a cell, we provide an equal chance for each pruned operation to be recovered through random sampling. When an operation is recovered, it replaces the zero operation in the super-net. To prevent a significant drop in performance caused by the insertion of useless operations, we initialize the recovered operations using the granularity parameters from before pruning, which are small enough. Consequently, only genuinely important operations will prevail in subsequent stages. The overall algorithm can be seen in Algorithm \ref{alg2}.

\section{Experiments}
In this section, we evaluate MG-DARTS on CIFAR-10, CIFAR-100 and ImageNet datasets, comparing it against various baseline methods in both DARTS-like and MobileNet-like search spaces. We try to answer the following research questions:
\begin{enumerate}
    \item RQ1: Can MG-DARTS discover an architecture with a better accuracy-efficiency trade-off compared with baselines? 
    \item RQ2: Can adaptive pruning consistently produce more appropriate retention ratios across different granularity levels for models with varying target sizes?
    \item RQ3: Can multi-stage search effectively reduce memory consumption without degrading search quality?
\end{enumerate}

Different variants for each proposed component in MG-DARTS is also discussed within the sub-sections.

\subsection{Datasets}
We evaluate our method on CIFAR-10, CIFAR-100 and ImageNet-1k, which are commonly used image classification datasets: 
\begin{enumerate}
    \item CIFAR-10: 10 classes of images with the resolution of $32\times32$. It has 50K training images and 10K test images. 
    \item CIFAR-100: 100 classes of images with the resolution of $32\times32$. It has 50K training images and 10K test images.
    \item ImageNet-1k: 1000 classes of images with the resolution of $224\times224$. It has 1.3M training images and 50K validation images. 
\end{enumerate}
Following the widely adopted ImageNet dataset setting \cite{xu2020pcdarts}, we use only 10\% training data during the search to reduce time.

\subsection{Implementation Details}
\paragraph{Search Space} We evaluate our method on the popular DARTS-like and MobileNet-like search space.

For DARTS search space, we follow the configurations of common settings \cite{liu2018darts,xu2020pcdarts,chen2019progressive}. There are two kinds of cells: Normal cells which maintain the resolution and channel number; Reduction cells which reduce the resolution by half and double the channel number. The reduction cells are set at the 1/3 and 2/3 of the depth of the architecture. The candidate operation set consists of 7 operations. And we remove most of the manual constraints on the search space as in \cite{bi2020gold}: Each edge can preserve an arbitrary number of operations, each node can preserve an arbitrary number of predecessors, and all the cells can have different structures. Moreover, we search for not only the operations, but also the filter numbers and weight sparsity patterns in convolution operations.

For MobileNet-like search space, we follow the configurations of SOTA works from the past five years \cite{wu2021stronger}. Since our method is primarily based on a typical differentiable architecture search pipeline, we currently cannot directly search for model depth. Therefore, we fix the depth at 17 layers, following the architecture discovered by WeakNAS \cite{wu2021stronger}. The search space consists of three operations: inverted residual blocks with kernel sizes of $3\times3$, $5\times5$, and $7\times7$. Instead of searching for the rigid expansion ratio, we directly optimize the filter numbers. Additionally, we explore weight sparsity patterns.

\paragraph{Hyper-parameters Settings} During the search, we decompose the super-net into three sub-nets according to the feature map resolution, i.e., $n$ is 3, where the later two sub-nets start with the two reduction cells. The regrowing ratio $r$ is set 0.2. In the architecture fine-tuning stage, we prune the network until reaching the target model size. To accelerate the search process at each stage, we initialize the thresholds $t_g$ using the minimum value of the current granularity parameters $g$, rather than starting from zero. We also apply the Cutout \cite{devries2017improved} and AutoAugment \cite{cubuk2019autoaugment} techniques, as well as the warm-up process of training the model weights for 5 epochs before the joint optimization with granularity parameters \cite{xu2020pcdarts,chen2019progressive}. The search process is conducted with a single NVIDIA A40 GPU. It takes around 1 day to search on both CIFAR-10 in DARTS search space and ImageNet in MobileNet-like search space.

\subsection{Baselines}
For the DARTS-like search space, most of existing methods are differentiable and only conduct operation-level search. Hence, we select 15 representative single-granularity methods \cite{liu2018darts,chen2019progressive,xu2020pcdarts,dong2019searching,bi2020gold,chen2021contrastive,yu2022cyclic,xiao2022shapley,ye2022b,xue2023improved,huang2023u,xiang2023zero,hu2024l,li2024adaptive,wei2025csc}, along with 2 latest multi-granularity methods \cite{mousavi2023dass,sukthanker2024weight} as baselines for CIFAR-10 and CIFAR-100 datasets, as shown in Table~\ref{tb1}. Among them, since DASS \cite{mousavi2023dass} is the most similar one to our method and is originally designed for extremely compact networks, we generate a small-size architecture using our approach and re-implement DASS under our customized search configuration to ensure a fair comparison. 

For the ImageNet dataset, we include only those baselines that report both model accuracy and model size, as summarized in Table~\ref{tb2}.

For the MobileNet-like search space, most of existing methods are non-differentiable and conduct operation-level and filter-level search. We compare with several state-of-the-art methods \cite{cai2019proxylessnas, chen2021contrastive, wu2021stronger, xiang2023zero}, as presented in Table~\ref{tb8}.

\begin{table*}
\begin{center}
\caption{Comparison with SOTA methods on CIFAR-10 and CIFAR-100 in DARTS-like search space.}
\label{tb1}
    \footnotesize
    \begin{tabular*}{\textwidth}{@{\extracolsep{\fill}}lcccccc}
       \toprule
       Method & \multicolumn{2}{c}{Top-1 Accuracy (\%)} & Params (M) & \multicolumn{2}{c}{\makecell{Accuracy Density\\(\%/M-params)}} & Granularity\\
       \cmidrule(lr){2-3} \cmidrule(lr){5-6}
        & CIFAR-10 & CIFAR-100 & & CIFAR-10 & CIFAR-100  \\ 
       \midrule
       DARTSV1 \cite{liu2018darts} & 97±0.14 & 82.24 & 3.3 & 29.39 & 24.92 & operation \\
       DARTSV2 \cite{liu2018darts} & 97.24 & 82.46 & 3.3 & 29.47 & 24.99 & operation \\
       P-DARTS \cite{chen2019progressive} & 97.5 & 83.45 & 3.4 & 28.68 & 24.54 & operation \\
       PC-DARTS \cite{xu2020pcdarts} & 97.43 & - & 3.6 & 27.06 & - & operation \\
       GDAS \cite{dong2019searching} & 97.07 & 81.62 & 3.4 & 28.55 & 24.01 & operation \\
       GOLD-NAS \cite{bi2020gold} & 97.47 & - & 3.67 & 26.56 & - & operation \\
       CTNAS \cite{chen2021contrastive} & 97.41 & - & 3.6 & 27.05 & - & operation \\
       CDARTS \cite{yu2022cyclic} & 97.52 & 84.31 & 3.9 & 25.01 & 21.62 & operation \\
       Shapley-NAS \cite{xiao2022shapley} & 97.53 & - & 3.6 & 27.09 & - & operation \\
       $\beta$-DARTS \cite{ye2022b} & 97.47 & 83.76 & 3.75 & 25.99 & 22.33 & operation \\
       PA-DARTS \cite{xue2023improved} & 97.52 & 83.46 & 3.57 & 27.31 & 23.71 & operation \\ 
       U-DARTS \cite{huang2023u} & 97.41 & 83.68 & 3.3 & 29.51 & 25.35 & operation \\
       Zero-Cost-PT \cite{xiang2023zero} & 97.38 & - & 4.6 & 21.16 & - & operation \\
       $l$-DARTS \cite{hu2024l} & 97.54 & 83.62 & 4.1 & 23.79 & 20.39 & operation \\
       ACA-PC-DARTS \cite{li2024adaptive} & 97.46 & - & 3.6 & 27.07 & - & operation \\
       CSC-DARTS \cite{wei2025csc} & 97.48 & 82.18 & 4.6 & 21.19 & 17.86 & operation \\
       TangleNAS \cite{sukthanker2024weight} & 97.44 & - & 3.58 & 27.21 & - & operation, filter \\
       MG-DARTS (ours, large) & 97.47 & 83.79 & 2.2 & \textbf{44.3} & \textbf{38.08} & operation, filter, weight \\
       \midrule
       DASS \cite{mousavi2023dass} (small) & 95.49 & 77.09 & 0.71 & 133.92 & 108.57 & operation, weight \\
       MG-DARTS (ours, small) & 96.84 & 81.76 & 0.7 & \textbf{138.34} & \textbf{116.8} & operation, filter, weight \\
       \bottomrule
    \end{tabular*}
    \end{center}
\end{table*}

\begin{table*}
\begin{center}
    \caption{Comparison with SOTA methods on ImageNet in DARTS-like search space.}
    \label{tb2}
    \footnotesize
    \begin{tabular*}{\textwidth}{@{\extracolsep{\fill}}lccccc}
       \toprule
        Method & \multicolumn{2}{c}{Accuracy (\%)} & Params (M) & \makecell{Accuracy Density\\(\%/M-params)} & Granularity\\
        \cmidrule(lr){2-3}
        & Top-1 & Top-5 &  \\ 
       \midrule
       DARTSV2 \cite{liu2018darts} & 73.3 & 91.3 & 4.7 & 15.6 & operation \\
       P-DARTS \cite{chen2019progressive} & 75.6 & 92.6 & 4.9 & 15.43 & operation \\
       PC-DARTS \cite{xu2020pcdarts} & 75.8 & 92.7 & 5.3 & 14.3 & operation \\
       GDAS \cite{dong2019searching} & 74 & 91.5 & 5.3 & 13.96 & operation \\
       GOLD-NAS \cite{bi2020gold} & 76.1 & 92.7 & 6.4 & 11.89 & operation \\
       CDARTS \cite{yu2022cyclic} & 75.9 & 92.6 & 5.4 & 14.06 & operation \\
       Shapley-NAS \cite{xiao2022shapley} & 76.1 & - & 5.4 & 14.09 & operation \\
       $\beta$-DARTS \cite{ye2022b} & 76.1 & 93 & 5.5 & 13.83 & operation \\
       PA-DARTS \cite{xue2023improved} & 75.3 & 92.25 & 5.2 & 14.48 & operation \\
       U-DARTS \cite{huang2023u} & 73.9 & 91.9 & 4.9 & 15.08 & operation \\
       Zero-Cost-PT \cite{xiang2023zero} & 75.4 & 92.4 & 6.3 & 11.96 & operation \\
       $l$-DARTS \cite{hu2024l} & 75.09 & 92.35 & 5.8 & 12.94 & operation \\
       CSC-DARTS \cite{wei2025csc} & 74.5 & 91.7 & 4.6 & 16.19 & operation \\
       TangleNAS \cite{sukthanker2024weight} & 74.3 & - & 5.15 & 14.42 & operation, filter \\
       MG-DARTS (ours, large) & 75.3 & 92 & 4 & \textbf{18.82} & operation, filter, weight \\
       \midrule
       DASS \cite{mousavi2023dass} (small) & 63.79 & 84.74 & 0.71 & 89.84 & operation, weight \\
       MG-DARTS (ours, small) & 64.89 & 86.36 & 0.7 & \textbf{92.7} & operation, filter, weight \\
       \bottomrule
    \end{tabular*}
\end{center}
\end{table*}

\begin{table}[ht]
\centering
\caption{Comparison with SOTA methods on ImageNet in MobileNet-like search space.}
\label{tb8}
\footnotesize
\setlength{\tabcolsep}{4pt} 
\begin{tabular}{lcccc}
\toprule
Method & \makecell{Top-1\\Accuracy\\(\%)} & \makecell{Params\\(M)} & \makecell{Accuracy\\Density\\(\%/M)} & Granularity \\
\midrule
CTNAS \cite{chen2021contrastive} & 77.3 & -- & -- & operation, filter \\
WeakNAS \cite{wu2021stronger} & 81.3 & 8.8 & 8.77 & operation, filter \\
Zero-Cost-PT \cite{xiang2023zero} & 76.4 & 8.0 & 9.55 & operation, filter \\ 
TangleNAS \cite{sukthanker2024weight} & 77.4 & 7.6 & 10.18 & operation, filter \\ 
\midrule
MG-DARTS & 77.2 & 7.4 & \textbf{10.43} & \makecell{operation, filter,\\weight} \\
\bottomrule
\end{tabular}
\end{table}

\begin{table}
\begin{center}
    \caption{Comparison of MG-DARTS in single-stage and multi-stage settings.}
    \label{tb5}
\footnotesize
    \begin{tabular}{ccccc}
       \toprule
       \multicolumn{2}{c}{Method} & \makecell{Maximum\\Memory\\Consumption (G)} & \makecell{Search\\Cost\\(h)} & \makecell{Accuracy\\(\%)}\\
       \midrule
       Single-stage &   & 26.4 & 8 & 96.56\\
       \midrule
       ~ & Stage 1 & 12.1 & \multirow{4}{*}{10} & \multirow{4}{*}{96.56}\\
       Multi-stage & Stage 2 & 12.4\\
       ~ & Stage 3 & 11.9\\
       ~ & Fine-tune & 7.8\\
       \bottomrule
    \end{tabular}
\end{center}
\end{table}

\begin{table}
\begin{center}
    \caption{Comparison of MG-DARTS with different regrowing ratios.}
    \label{tb4}
\footnotesize
\setlength{\tabcolsep}{8pt} 
    \begin{tabular}{cccccc}
       \toprule
       \makecell{Regrowing \\Ratio $r$} & 
       \makecell{Top-1 \\Accuracy (\%)} & \makecell{Params (M)} & 
       \makecell{Accuracy Density \\(\%/M-params)}\\
       \midrule
       0 & 97.14 & 2.4 & 40.47\\
       0.1 & 97.21 & 2.4 & 40.5\\
       0.2 & 97.47 & 2.2 & 44.30\\
       \bottomrule
    \end{tabular}
\end{center}
\end{table}

\begin{table*}[]
\begin{center}
    \caption{Comparison with SOTA methods in different DARTS-like search spaces on CIFAR10.}
    \label{tb3}
    \footnotesize
    \setlength{\tabcolsep}{8pt} 
    \begin{tabular}{lccccccc}
        \toprule
            \multirow{2}{*}{Method} & \multicolumn{2}{c}{Search Space 1} & \multicolumn{2}{c}{Search Space 2} & \multicolumn{2}{c}{Search Space 3} & \multirow{2}{*}{\makecell{Memory Reduction\\Strategy}}\\
            \cmidrule(lr){2-3}\cmidrule(lr){4-5}\cmidrule(lr){6-7}
            & \makecell{Top-1 Accuracy\\(\%)} &  Params (M) & \makecell{Top-1 Accuracy\\(\%)} &  Params (M) & \makecell{Top-1 Accuracy\\(\%)} &  Params (M) \\ 
            \midrule
            DARTSV1 \cite{liu2018darts} & 96.85 & 1.23 & 96.69 & 1.37 & - & - & Depth-level \\
            PC-DARTS \cite{xu2020pcdarts} & 97.06 & 1.68 & 96.66 & 1.47 & 96.56 & 1.47 & Width-level \\
            GDAS \cite{dong2019searching} & 96.78 & 0.95 & 95.34 & 0.65 & 96.11 & 1.13 & Path-level \\ \midrule
            MG-DARTS & 97.04 & 1.34 & 97.02 & 1.25 & 97.10 & 1.19 & \makecell{Depth-level\\with Re-evaluation} \\
        \bottomrule
    \end{tabular}
    \end{center}
\end{table*}

\subsection{MG-DARTS Discovers Architectures with a Better Accuracy-Efficiency Trade-Off (RQ1)}
In this section, we verify whether MG-DARTS can discover architectures with a better accuracy-efficiency trade-off compared with baselines.
\subsubsection{Results on CIFAR-10 and CIFAR-100 in DARTS search space}
We implement MG-DARTS with two different search space settings. In order to compare with most of single-granularity methods that prioritize accuracy, we adopt a large search space configuration, which includes relatively large models with 14 cells and an initial filter number of 44. On the other hand, to compare with most of multi-granularity methods that prioritize model efficiency and require higher search cost, we adopt a small search space configuration, exploring more compact models with 8 cells and an initial filter number of 24. For fair comparison, we adopt the same depth and initial width settings for the baseline DASS. We set the weight retention ratio to 90\% for DASS during pruning, as this ratio shows the most promising results under our search configuration according to our toy ablation study (Table~\ref{tb6}). The results are shown in Table \ref{tb1}. We utilize accuracy density \cite{bianco2018benchmark} to measure the accuracy-efficiency trade-off, which quantifies how efficiently each model uses its parameters by dividing top-1 accuracy by parameter number.

In the large model setting, MG-DARTS achieves comparable accuracy of around 97.5\% and 83.8\% on CIFAR-10 and CIFAR-100, respectively, using only 2.2M parameters. On CIFAR-10, MG-DARTS reduces model size by an average of up to 40\% compared to baselines with a marginal accuracy difference of 0.05\%. For CIFAR-100, MG-DARTS outperforms all baselines in accuracy except CDARTS, while achieving up to a 43\% reduction in model size compared to CDARTS, with only a modest accuracy gap of around 0.5\%. Additionally, MG-DARTS outperforms all the baselines in terms of the accuracy density, showcasing the advantages of exploring a comprehensive multi-granularity search space. 

In the small model setting, MG-DARTS exhibits a higher accuracy than DASS while maintaining a similar parameter count around 0.7M. 

It is worth mentioning that in the original DASS work, an impressive 95.31\% accuracy is achieved on CIFAR-10 with only 0.1M parameters. However, under our 0.7M parameter setting, DASS only reaches 95.49\% accuracy, suggesting that it may not be fully leveraging its realizing under our experimental configuration.

We attribute this discrepancy to two main factors. First, we adopt the same pretraining schedule as reported in the original paper, i.e. 50 epochs of super-net training and 20 epochs of pruning, which was designed for a smaller 16-channel super-net. In our 24-channel setting, the super-net likely requires longer training to converge, potentially leading to suboptimal pruning masks. Second, to ensure a fair comparison with our method, we retrain all discovered architectures from scratch, whereas original DASS inherits pretrained weights to restore accuracy during evaluation.

\begin{table}[ht]
\centering
\footnotesize
\setlength{\tabcolsep}{3pt} 
\caption{Comparison of DASS and MG-DARTS under different search and evaluation settings. MG-DARTS is less sensitive to the degree of super-net optimization.}
\label{tb6}
\begin{tabularx}{\linewidth}{lccccc}
\toprule
Method & \makecell{Search\\Epoch} & \makecell{Evaluation\\Mode} & \makecell{Weight Ratio\\(\%)} & \makecell{Accuracy\\(\%)} & \makecell{Search Cost\\(h)} \\
\midrule
\multirow{16}{*}{DASS} & \multirow{10}{*}{10} & \multirow{6}{*}{\makecell{from\\scratch}} 
    & 30    & 73.44 & \multirow{8}{*}{7} \\
&   &           & 50    & 80.36 &   \\
&   &           & 70    & 84.68 &   \\
&   &           & 90    & 86.58 &   \\
&   &           & 96.3  & 86.82 &   \\
&   &           & 100   & 86.35 &   \\
\cmidrule{3-5}
&   & \multirow{4}{*}{finetune} 
& 50    & 79.62 &   \\
&   &           & 90    & 86.43 &   \\
&   &           & 96.3  & 86.52 &   \\
&   &           & 100   & 86.75 &   \\
\cmidrule{2-6}
& \multirow{6}{*}{70} & \multirow{3}{*}{\makecell{from\\scratch}}     
& 50    & 86.30 & \multirow{6}{*}{55} \\
&   &           & 90    & 87.01 &   \\
&   &           & 100   & 86.66 &   \\
\cmidrule{3-5}
&   & \multirow{3}{*}{finetune} 
& 50    & 87.12 &   \\
&   &           & 90    & 87.78 &   \\
&   &           & 100   & 87.57 &   \\
\midrule
MG-DARTS & 10 & \makecell{from\\scratch} & Adaptive (96.3) & 93.37 & 5 \\
\bottomrule
\end{tabularx}
\end{table}

To support this analysis, we conduct controlled experiments on CIFAR-10 using DASS under various search and evaluation settings. We vary the weight retention ratio (30\%, 50\%, 70\%, and 90\%) while adjusting the number of operations to keep the overall model size approximately constant around 0.7M. Additionally, we compare different numbers of search epochs (10 and 70) to simulate varying levels of super-net optimization. For evaluation, we consider both training from scratch and fine-tuning based on the pretrained super-net weights.

As shown in Table~\ref{tb6}, DASS is highly sensitive to the retention ratio when the super-net is insufficiently optimized. With only 10 search epochs, models with lower retention ratios (e.g., 30\% and 50\%) suffer significant performance drops, likely due to suboptimal sparsity patterns. Fine-tuning based on pretrained weights during evaluation partially mitigates this issue in sparse regimes (e.g., a 6.18\% improvement at 50\% ratio). Increasing the search epochs to 70 also alleviates this problem, particularly under high sparsity levels (e.g., a 5.94\% gain at 50\% ratio). These results confirm DASS’s reliance on pretraining.

In contrast, MG-DARTS achieves promising result when trained from scratch even without extensive pretraining. To eliminate the potential influence of search space topology constraints for fair comparison, we also apply the adaptive retention ratio obtained by MG-DARTS to DASS. This yields the best scratch-trained result at 86.82\% among all DASS settings, and performs only slightly worse than the fully dense model when fine-tuned. These results suggest that MG-DARTS, by adaptively determining retention ratios at multiple granularity levels, is more robust to under-optimized super-nets and inaccurate architectural patterns.

Overall, the performance gap reflects two fundamentally different design philosophies: DASS emphasizes aggressive sparsity through heavy pretraining, whereas MG-DARTS promotes resilience by adaptively controlling retention across multiple granularity levels.

\subsubsection{Results on ImageNet in DARTS search space}
Similarly, we employ two different settings for ImageNet. For the large model setting, we explore models with 14 cells and an initial filter number of 48. For the small model setting, given the high search cost of the baseline DASS, we directly transfer the small-size architectures discovered on CIFAR-10. The results are provided in Table \ref{tb2}. 

As shown in the table, MG-DARTS surpasses all the baselines in terms of accuracy density, which indicates that MG-DARTS can be generalized well to large datasets. The transfered architecture discovered by MG-DARTS also demonstrates higher accuracy compared to that discovered by DASS. This highlights the strong generalization capability of the architectures generated by our method.

Additionally, following GOLD-NAS \cite{bi2020gold}, our method adopts the basic one-level relaxation of the DARTS optimization strategy without incorporating any regularization techniques. This design choice was made to ensure simplicity and clearly highlight the core contributions of our work. It is worth noting that many well-established regularization techniques, such as $\beta$-DARTS \cite{ye2022b}, are orthogonal to our framework and can be seamlessly integrated to further improve performance.

\subsubsection{Results on ImageNet in MobileNet-like search space}
To evaluate the effectiveness of our method across different search spaces, we also implement it in a MobileNet-like search space. As shown in Table \ref{tb8}, MG-DARTS achieves comparable top-1 accuracy on ImageNet while reducing model size by incorporating weight-level granularity and determining appropriate retention ratios through joint discretization.

\subsection{Adaptive Pruning Consistently Achieves a Better Granularity Balance at Different Model Sizes (RQ2)}
The superiority of MG-DARTS over methods with manually defined ratios has already been demonstrated for certain model sizes. In this section, we verify whether adaptive pruning can consistently achieve a better granularity balance for different target model sizes. To eliminate the effect of multi-stage search, we employ a smaller search space setting with 8 cells, where the constructed super-net is small enough to be directly put into the memory without decomposition. 

In this section, the retention ratio is defined as the ratio of the remaining unit number to the original unit number. For operations, the original unit number refers to the initial operation number in the super-net. For filters, the original unit number denotes the initial filter number in operations within the discovered architecture. For weights, the original unit number denotes the initial weight number in filters within the discovered architecture. 

\subsubsection{Comparison with different retention ratios}
\begin{figure}[t]
\centering
\includegraphics[width=0.9\columnwidth]{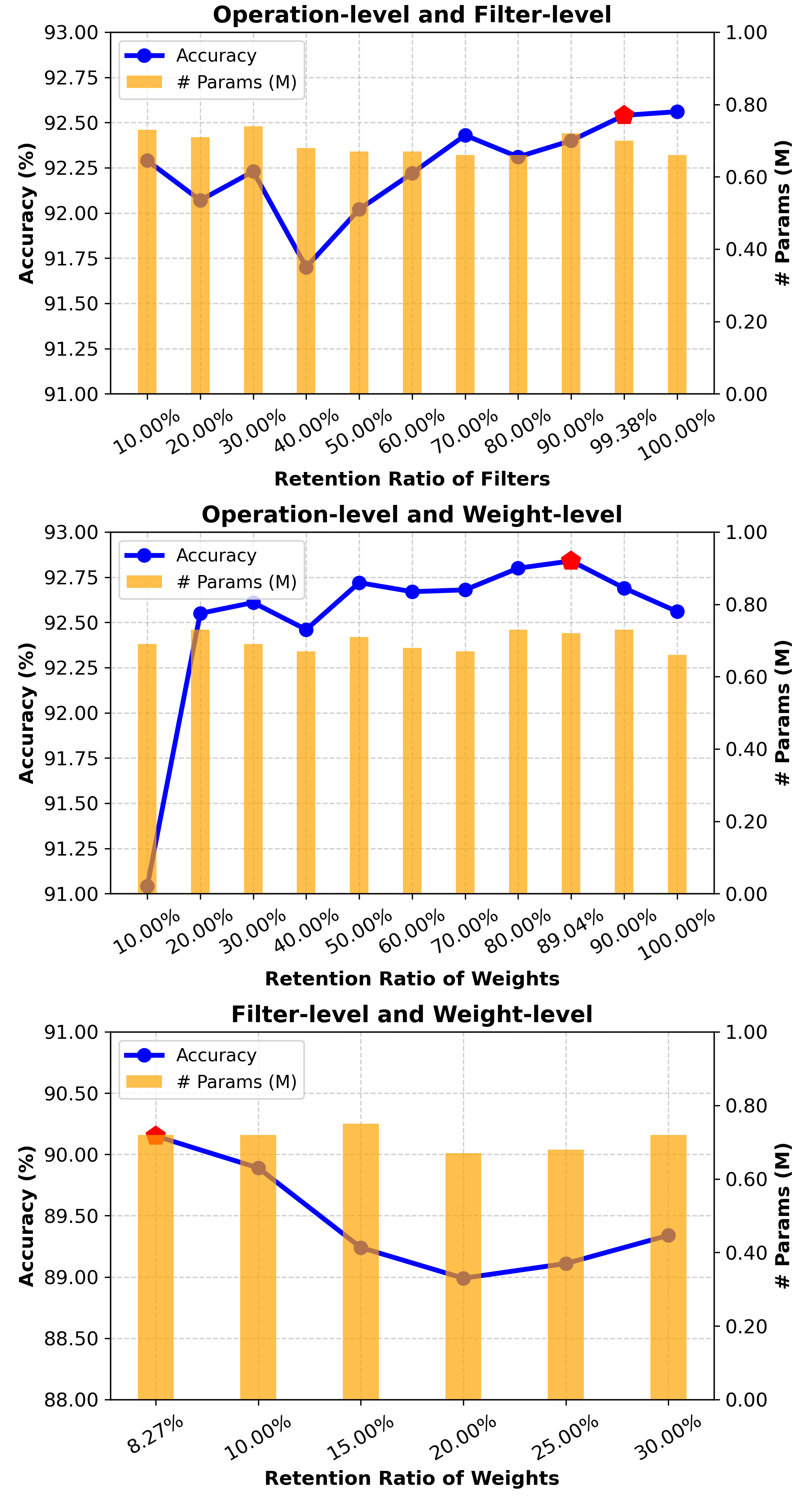}
\caption{Comparison with different manual ratios. Adaptive pruning effectively identifies configurations that lie within high-performance region.}
\label{fig6}
\end{figure}

In this section, we conduct the search using combinations of any two granularity levels and manually vary the retention ratios of units at each level to evaluate whether our adaptive pruning method effectively determines appropriate ratios. The results are shown in Fig. \ref{fig6}, where the ratios discovered by our method are marked with a star.

For the operation-filter combination, we increase the filter retention ratio by 10\% at each step and adjust the operation count accordingly to maintain a similar overall model size. Similarly, for the operation–weight and filter–weight combinations, we increase the weight retention ratio by 10\% and 5\% respectively, while adjusting the corresponding filter or operation count. This setup allows us to verify whether the discovered ratios fall within high-performance regions.

Note that the number of filters found during the search may slightly differ from those in the final architecture due to post-search filter alignment. To ensure consistency, we report the retention ratios obtained directly from the search phase. While the exact numerical ratios may not always match, the relative proportions of filters across different settings remain consistent, ensuring a fair comparison.

As shown in the figure, the optimal retention ratio varies across different granularity combinations, with performance peaks appearing at several points. Nevertheless, adaptive pruning effectively identifies configurations that lie close to these peaks and yields top-tier performance across all settings.

\subsubsection{Comparison with different pruning strategies}
To simplify the process, we only conduct search at the operation and filter levels. We compare our method with the widely used progressive pruning strategy that relies on predefined pruning rates. Through a single search process, our method generates a series of architectures with varying sizes using adaptive pruning. We then calculate the average pruning rates at both the operation and filter levels, which are approximately 36 operations and 0.7\% filters pruned every two epochs.

Based on these observed rates, we design two baseline methods to serve as comparisons. This ensures that the overall retention ratios of the baselines are reasonably close to our method, so that our advantages can be better highlighted. Specifically, baseline 1 adopts a higher pruning rate at the operation level and a lower rate at the filter level, while baseline 2 adopts a higher filter-level pruning rate and a lower operation-level pruning rate:
\begin{enumerate}
    \item Fixed pruning rates 1: 44 operations and 0.3\% filters per two epochs
    \item Fixed pruning rates 2: 28 operations and 1.5\% filters per two epochs
\end{enumerate}
Employing these baseline methods, we obtain baseline architectures that exhibit variations in the retention ratios at operation and filter levels.  

\begin{figure}[t]
\centering
\includegraphics[width=1\columnwidth]{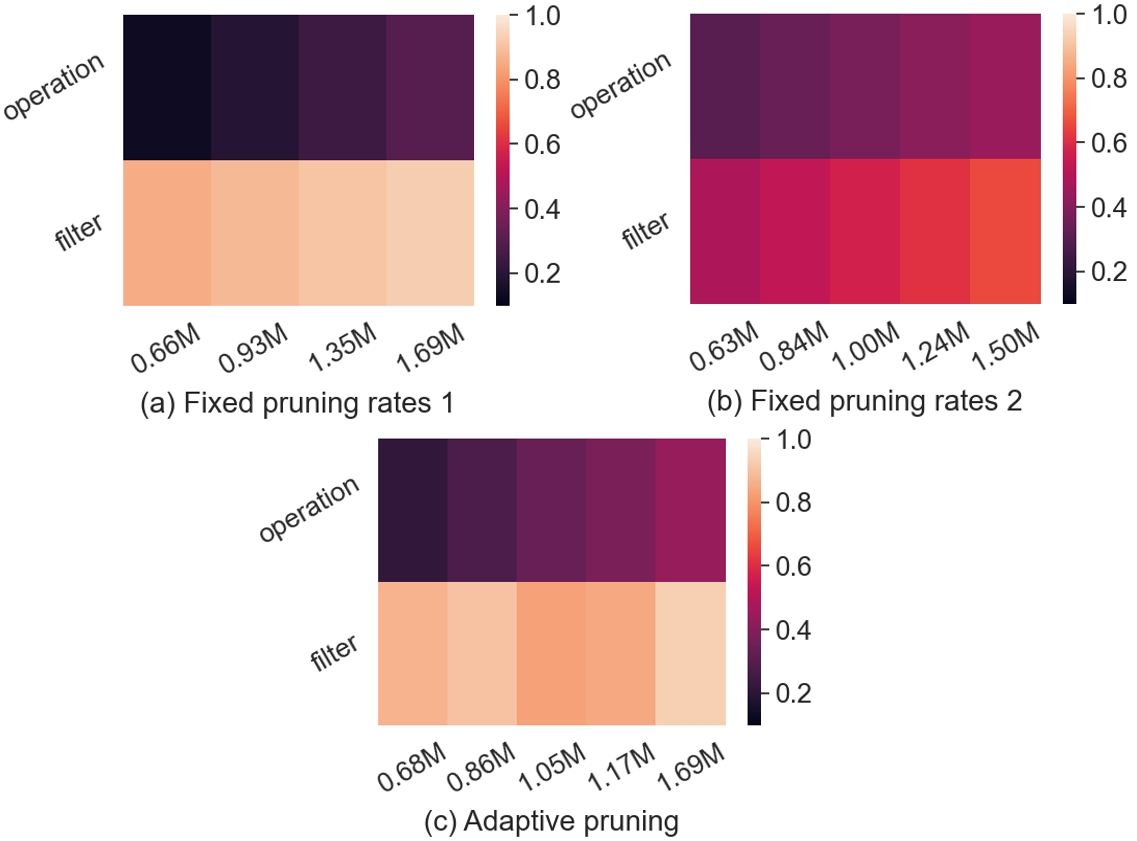} 
\caption{Retention ratios of operations and filters in pruning with fixed pruning rates and our proposed adaptive pruning.}
\label{fig7}
\end{figure}

The retention ratios in the discovered architectures of adaptive pruning and the baselines are depicted in Fig. \ref{fig7}. As expected, baseline 1 generally retains more filters and fewer operations compared to adaptive pruning at similar model sizes. Conversely, baseline 2 preserves more operations and fewer filters. Moreover, it can be observed that adaptive pruning generates slimmer architectures at sizes of 1.05M and 1.17M, while producing wider architectures at smaller model sizes. This illustrates its ability to adaptively adjust the retention ratios of operations and filters as the architecture evolves.

\begin{figure}[t]
\centering
\includegraphics[width=0.8\columnwidth]{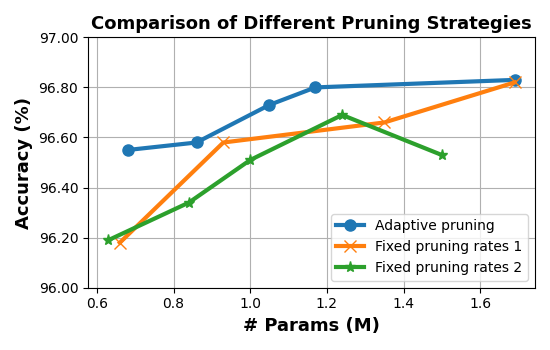}
\caption{Comparison with progressive pruning with fixed pruning rates. Adaptive pruning consistently yields models with improved performance.}
\label{fig8}
\end{figure}

The performance of these methods is illustrated in Fig. \ref{fig8}. As shown in the figure, adaptive pruning consistently outperforms both baselines across all model sizes, forming a Pareto frontier. This indicates its effectiveness in dynamically determining appropriate retention ratios for different target complexities.

\subsubsection{Ablation of different granularity levels}
To evaluate the importance of each granularity in our method, we conduct experiments comparing architectures searched with different combinations of granularity levels. As shown in Table \ref{tb9}, incorporating filter-level and weight-level searchable units leads to improved model performance at similar model size, as redundancy is better identified and reduced. The best performance is achieved when all three levels are jointly considered.

\begin{table}[ht]
\centering
\caption{Performance Comparison of search spaces with different granularity levels.}
\label{tb9}
\footnotesize
\setlength{\tabcolsep}{12pt} 
\begin{tabular}{lcc}
\toprule
Search Space & Accuracy (\%) & Params (M) \\
\midrule
operation & 93.11 & 0.7 \\
operation + filter & 93.21 & 0.71 \\
operation + weight & 93.37 & 0.72 \\
operation + filter + weight & 93.52 & 0.71 \\
\bottomrule
\end{tabular}
\end{table}

\begin{table}[ht]
\centering
\caption{Performance Comparison of different optimization strategies.}
\label{tb10}
\footnotesize
\setlength{\tabcolsep}{16pt} 
\begin{tabular}{lcc}
\toprule
Strategy & Accuracy (\%) & Params (M) \\
\midrule
Iterative (Full Epoch) & 97.13 & 2 \\
Iterative (Half Epoch) & 97.19 & 2.2 \\
Joint & 97.47 & 2.2 \\
\bottomrule
\end{tabular}
\end{table}

\subsubsection{Effect of the optimization strategy}
To optimize the granularity parameters $g$ (i.e., $\alpha, \beta, \omega$) and their associated thresholds $t_g$ (i.e., $t_\alpha, t_\beta, t_\omega$), a natural idea is to iteratively optimize $g$ and $t_g$ in an alternating manner to avoid conflicting gradient signals. However, as shown in Table~\ref{tb10}, we empirically find that the joint optimization strategy, where both $g$ and $t_g$ are updated simultaneously, achieves the best performance. It obtains the highest accuracy of 97.47\%, outperforming the iterative strategies with half-epoch (97.19\%) and full-epoch (97.13\%) update intervals.

We attribute the inferior performance of iterative strategies to two main issues: error propagation and reduced update frequency. Specifically, iterative updates are designed to approximate the following bi-level optimization problem:
\begin{equation}
\begin{aligned}
\min_{t_g} \ \mathcal{L}(g^*(t_g), t_g), \\
\text{where} \quad g^*(t_g) = \arg\min_{g} \ \mathcal{L}(g, t_g)
\end{aligned}
\label{eq9}
\end{equation}
Since $g^*(t_g)$ cannot be solved exactly, iterative methods rely on partial updates of $g$, which results in inaccurate gradients for $t_g$  \cite{bi2019stabilizing}, leading to error propagation. Moreover, as only one parameter group is updated at a time, both $g$ and $t_g$ receive only half as many updates as they would under joint optimization under the same training budget, which slows down convergence and degrades the performance.

\subsection{Multi-Stage Search Ensures Low Memory Consumption with High Search Quality (RQ3)}
In this section, we aim to assess the impact of multi-stage search on reducing memory consumption while maintaining high search quality.

\subsubsection{Reduction of memory consumption}
To evaluate the reduction in memory consumption, we conduct a comparison between the search processes using both single-stage and multi-stage settings on CIFAR-10 under the small search space configuration. The results are presented in Table \ref{tb5}. 

As shown in the table, the peak memory consumption in the multi-stage search (12.4G) is only 46.9\% of that in the single-stage counterpart (26.4G), with merely 2 additional GPU hours incurred and no accuracy drop observed. This indicates that the multi-stage design effectively reduces memory usage at a modest computational cost, without compromising performance.

\subsubsection{Preservation of search quality}
To assess whether our multi-stage search can consistently preserve high search quality, we compare it with other memory reduction strategies in different variations of the search space. For simplicity, only operation-level search is conducted. 

We choose DARTSV1, PC-DARTS and GDAS as baseline methods to represent depth-level, width-level and path-level strategies, respectively. We evaluate these methods on three 8-cell search spaces with an initial filter number of 36. Search space 1 represents the original DARTS search space. Search space 2 relaxes the connection constraint, allowing for the discovery of multi-path architectures. Search space 3 further relaxes the identical cell constraint, enabling different cell structures. The results are shown in Table \ref{tb3}.

In search space 1, multi-stage search outperforms both DARTSV1 and GDAS in terms of accuracy and achieves similar performance with PC-DARTS, highlighting its effectiveness in preserving high search quality. After removing the manual constraints on the search spaces, the accuracy of the architectures discovered by baseline methods decrease significantly. Specifically, the path-level method GDAS performs poorly in Search Space 2, indicating its limited ability to explore multi-path architectures. Similarly, PC-DARTS, which performs well in search space 1, experiences an obvious accuracy drop in search spaces 2 and 3, revealing the width-level strategy's failure in preserving search quality when the search space exhibits more flexibility. In comparison, our multi-stage search maintains high search quality in all the three search spaces.

\subsubsection{Effect of hyper-parameter $r$}
To assess the significance of the regrowing process and the effect of the hyper-parameter $r$, we conduct experiments comparing various regrowing ratios on CIFAR-10 in the large search space. The results are presented in Table \ref{tb4}. 

From the table, it is apparent that the absence of the regrowing process, represented by $r$ equaling 0, leads to limited performance. This is reasonable since units that have been pre-maturely pruned cannot be recovered. As the value of $r$ increases, there is a corresponding improvement in accuracy from 97.14\% to 97.47\%. 

Note that the increase of $r$ will also add to the memory consumption and search cost. To avoid excessive burden on memory and time, we choose $r=0.2$ as the final setting and opt not to further increase its value.

\subsubsection{Effect of intermediate task-specific auxiliary layers}
To investigate the impact of intermediate auxiliary layers in the multi-stage search strategy, we compare three different configurations, all of which primarily consist of two convolutional layers followed by a dense layer as the classifier. Our design principle is to align the feature map size with the final output while keeping the auxiliary layers lightweight, ensuring that the primary focus of optimization remains on the search space. Their size is kept around 10\% of the main part of the search space, ensuring that the total parameter count, including both the discovered architecture at the current stage and its following auxiliary layers, remains approximately the same as that of the final discovered architecture.

We denote the configuration used in our method as the medium size. Based on this, we create two variants: the large size, where we increase the kernel size of the convolutional layers, and the small size, where we reduce the number of channels in the convolutional layers. To evaluate the performance under these different configurations, we train the discovered architectures for 200 epochs.

As shown in Table \ref{tb7}, the auxiliary task-specific layers need to be within a reasonable size range to ensure good performance. Their primary purpose is to simulate high-level feature extraction and prediction for the earlier stages, allowing the early-stage layers to focus on capturing low-level features. If these auxiliary layers are too small, the final layers of the early-stage sub-nets may be forced to evolve into high-level feature extractors themselves, which disrupts the intended division of labor and degrades overall performance.

\begin{table}[ht]
\centering
\caption{Performance comparison of different auxiliary layers.}
\label{tb7}
\footnotesize
\setlength{\tabcolsep}{20pt} 
\begin{tabular}{lcc}
\toprule
Component & Params (M) & Accuracy (\%) \\
\midrule
Small Aux & 0.69 & 94.26 \\
Medium Aux & 2.69 & 94.67 \\
Large Aux & 3.23 & 94.76 \\
\midrule
Main Parts & 21 & - \\
\bottomrule
\end{tabular}
\end{table}

\begin{figure}[t]
\centering
\includegraphics[width=0.9\columnwidth]{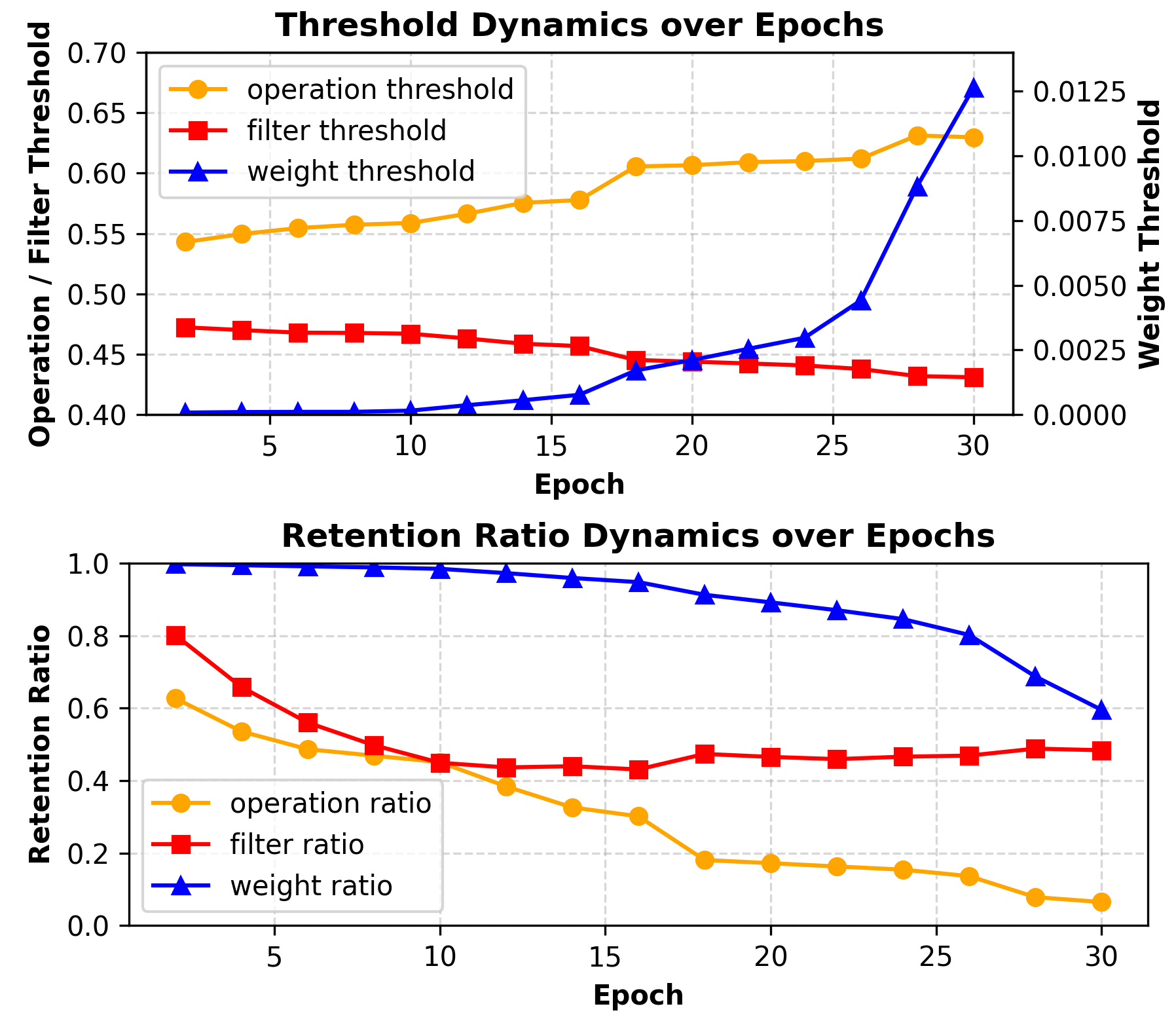}
\caption{Dynamics of thresholds and retention ratios at different granularity levels during search.}
\label{fig9}
\end{figure}

\begin{figure}[t]
\centering
\includegraphics[width=1\columnwidth]{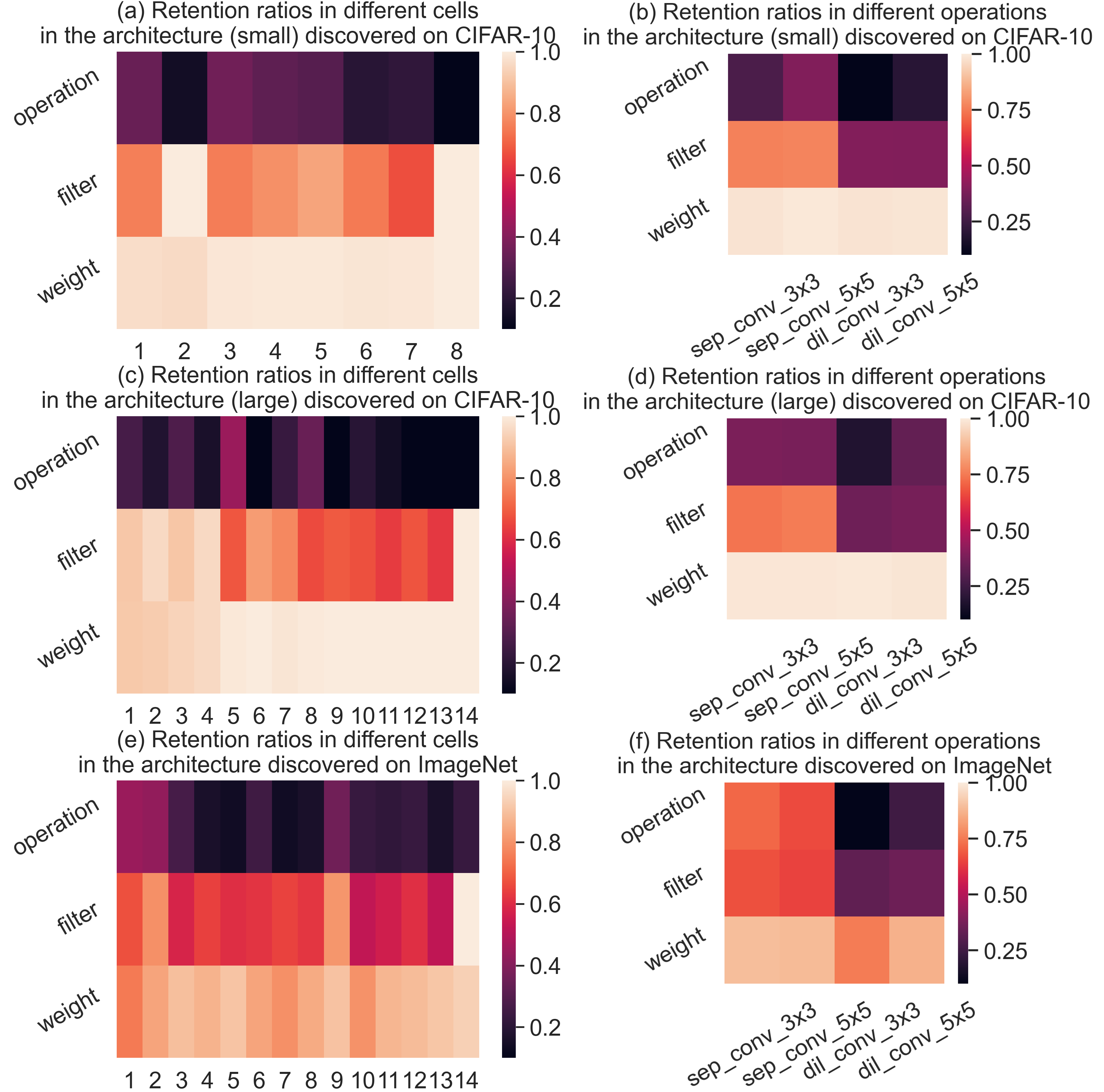} 
\caption{Distribution patterns of local units in discovered architectures. This figure illustrates the variance in unit distribution across different cells and operation types within the neural network architectures discovered on CIFAR-10 and ImageNet datasets. Notably, it highlights a tendency to preserve more operations in the early layers and more filters at both ends, as well as a general preference for separable over dilated convolution operations.}
\label{fig4}
\end{figure}

\subsection{Visualization}
To further understand the proposed MG-DARTS search process and the resulting architectures, we provide two visual analysis. The first part illustrates the evolution of the architectures throughout the search process, while the second part presents the architectural patterns of the discovered architectures.

\subsubsection{Architectural Evolution during Search}
Fig.~\ref{fig9} visualizes the epoch-wise evolution of architectural parameters during the search process under the small search space configuration. The upper plot depicts the progression of the average learned thresholds, while the lower plot presents the corresponding retention ratios.

The thresholds for operation and filter levels exhibit a relatively smooth and gradual variation, whereas the weight-level threshold remains low for early epochs and rises sharply later, indicating delayed pruning at the weight level. From the retention ratio curves, we can observe a clearer coarse-to-fine pruning pattern, where the operation and filter ratios decrease quickly during the early epochs, while the weight ratio stays close to 1.0 initially and only begins to decline significantly in the later phase. 

This behavior can be attributed to the inherent stability differences among parameters at different granularity levels. In the early phase, the super-net is under-optimized and subject to noisy gradient signals. Coarser-grained structures, due to their limited candidate space and larger influence on performance, tend to exhibit more consistent and distinguishable importance even under such unstable conditions. In contrast, finer-grained structures are more sensitive to noise, causing their thresholds to receive less consistent gradients and update more slowly. This inherent coarse-to-fine pruning pattern naturally contributes to a more stable and reliable search process.

\subsubsection{Architectural Patterns in Discovered Architectures}
In this section, we investigate the local unit distribution of convolution operations within the discovered architectures from Table~\ref{tb1} and \ref{tb2} to uncover general structural patterns. Figure~\ref{fig4} presents the retention ratios of units at different granularity levels, organized by layer depth and operation type.

At the layer level, more operations are kept in the shallower part of the network, which is reasonable since these regions are responsible for processing raw input features and benefit from diverse transformation pathways to extract low-level information. At the filter level, a U-shaped distribution is observed: fewer filters are maintained in the middle, while both ends retain more. This pattern is consistent with previous pruning studies \cite{meng2020pruning}, which suggest that filter redundancy tends to be higher in the middle part of the network. Correspondingly, more weights are preserved in the middle, partially due to the reduced number of filters, resulting in denser parameter allocation per filter to maintain sufficient expressiveness.

At the operation-type level, separable convolutions are consistently preferred over dilated ones, in line with findings from most prior differentiable architecture search works \cite{liu2018darts,xu2020pcdarts}. This preference is also reflected in the greater allocation of filters and weights to separable convolutions, highlighting their central role in the resulting architectures.

\section{Conclusion}
In this work, we introduce a new perspective to discover effective yet parameter-efficient neural networks from scratch by comprehensively exploring searchable units of multiple granularities and adaptively adjusting the retention ratios of the multi-granularity units. Additionally, we effectively tackle the challenge of high memory consumption of the search process without compromising the search quality via depth-level decomposition with progressive re-evaluation. Experimental results demonstrate the consistent superiority of MG-DARTS over other alternatives in terms of the accuracy-parameter trade-off. To further reduce the search cost, our future research efforts will focus on integrating zero-cost metrics to estimate the importance of searchable units at initialization.

\section{Acknowledgments}
\noindent This work was supported in part by the Hong Kong RGC General Research Fund (No.: PolyU 15205924); and in part by the Research Institute for Artificial Intelligence of Things, The Hong Kong Polytechnic University.

\bibliography{Submission_Final}

\bibliographystyle{IEEEtran}

\newpage

\vfill

\end{document}